\documentclass[10pt,twocolumn,letterpaper]{article}

\usepackage[pagenumbers]{cvpr} 

\definecolor{cvprblue}{rgb}{0.21,0.49,0.74}
\usepackage[accsupp]{axessibility}
\usepackage[pagebackref,breaklinks,colorlinks,allcolors=cvprblue]{hyperref}
\usepackage{adjustbox}
\usepackage{url}            
\usepackage{booktabs}       
\usepackage{amsfonts}       
\usepackage{nicefrac}       
\usepackage{microtype}      
\usepackage{xcolor}         
\usepackage{enumitem}
\usepackage{bm}
\usepackage{wrapfig}
\usepackage{amsmath}
\usepackage{multicol}
\usepackage{multirow}
\usepackage{subcaption}
\usepackage{xcolor,colortbl}
\usepackage{dsfont}

\DeclareMathOperator*{\varr}{var}

\definecolor{yellow}{rgb}{1, 1, 0.7}
\definecolor{orange}{rgb}{1, 0.85, 0.7}
\definecolor{red}{rgb}{1, 0.7, 0.7}
\definecolor{normalred}{rgb}{1, 0, 0}

\def\paperID{9045} 
\def\confName{CVPR}
\def\confYear{2025}

\title{NeRFPrior: Learning Neural Radiance Field as a Prior for Indoor Scene Reconstruction}

\author{
Wenyuan Zhang$^1$, Emily Yue-ting Jia$^1$, Junsheng Zhou$^1$, Baorui Ma$^1$, Kanle Shi$^2$, \\
Yu-Shen Liu$^1$\thanks{The corresponding author is Yu-Shen Liu. This work was partially supported by Deep Earth Probe and Mineral Resources Exploration\text{—}National Science and Technology Major Project (2024ZD1003405), and the National Natural Science Foundation of China (62272263).} \ , Zhizhong Han$^3$    \\
  School of Software, Tsinghua University, Beijing, China$^1$ \\
  Kuaishou Technology, Beijing, China$^2$ \\
  Department of Computer Science, Wayne State University, Detroit, USA$^3$ \\
  {\tt\small zhangwen21@mails.tsinghua.edu.cn, jiaemily120@gmail.com, zhou-js24@mails.tsinghua.edu.cn} \\
  {\tt\small mabaorui2014@gmail.com, shikanle@kuaishou.com, liuyushen@tsinghua.edu.cn, h312h@wayne.edu} \\
}

\begin{document}
\maketitle

\begin{abstract}
    Recently, it has shown that priors are vital for neural implicit functions to reconstruct high-quality surfaces from multi-view RGB images. However, current priors require large-scale pre-training, and merely provide geometric clues without considering the importance of color. In this paper, we present NeRFPrior, which adopts a neural radiance field as a prior to learn signed distance fields using volume rendering for surface reconstruction. Our NeRF prior can provide both geometric and color clues, and also get trained fast under the same scene without additional data. Based on the NeRF prior, we are enabled to learn a signed distance function (SDF) by explicitly imposing a multi-view consistency constraint on each ray intersection for surface inference. Specifically, at each ray intersection, we use the density in the prior as a coarse geometry estimation, while using the color near the surface as a clue to check its visibility from another view angle. For the textureless areas where the multi-view consistency constraint does not work well, we further introduce a depth consistency loss with confidence weights to infer the SDF. Our experimental results outperform the state-of-the-art methods under the widely used benchmarks. Project page: \url{https://wen-yuan-zhang.github.io/NeRFPrior/}.
    
\end{abstract}

\section{Introduction}

3D surface reconstruction from multi-view images is a long-standing challenge in computer vision and graphics. Traditional methods, like multi-view-stereo (MVS) ~\cite{schonberger2016structure, yao2018mvsnet, goesele2006multi}, estimate 3D geometry by first extracting a sparse point cloud and then applying dense reconstruction on it. 
The latest reconstruction methods~\cite{wang2021neus, oechsle2021unisurf, yariv2021volsdf} learn implicit functions from multiple images via volume rendering using neural networks. These methods require learning priors~\cite{guo2022neural, yu2022monosdf, zhang2022regsdf} from an additional large-scale dataset to reveal accurate geometry and structure. However, these data-driven priors do not generalize well to other kinds of scenes which are different from the pretrained datasets, which drastically degenerates the performance.

Instead, some methods~\cite{fu2022geoneus, deng2022dsnerf, chen2021mvsnerf} introduce overfitting based priors to improve the generalization, since these priors can be learned by directly overfitting a single scene. Methods like MVS are widely adopted to extract overfitting priors, which use the photometric consistency to overfit a scene. However, these priors can merely provide geometric information and do not provide photometric information which is important for the network to predict colors in volume rendering.

To address this issue, we propose NeRFPrior, which introduces a neural radiance field as a prior to learn signed distance functions (SDF) to reconstruct smooth and high-quality surfaces from multi-view images. Thanks for current advanced training techniques for radiance fields~\cite{fridovich2022plenoxels, muller2022instant, chen2022tensorf, sun2022direct,kerbl20233dgs}, we are able to train a radiance field by overfitting multi-view images of a scene in minutes. Although more recent 3DGS methods~\cite{kerbl20233dgs} present a very promising solution for learning radiance fields with explicit 3D Gaussians, it is still a challenge to recover continuous SDFs from discrete, scattered, or even sparse 3D Gaussians. Per this, we adopt NeRF and leverage the trained NeRF as a prior to provide the geometry and color information of the scene itself. This enables us to learn a more precise SDF by explicitly imposing a multi-view consistency constraint on each ray intersection for its SDF inference. 

Specifically, to get the prior geometry, we query the density from the NeRF prior as an additional supervision for our neural implicit networks. With the predicted density at each sample along a ray, we find the intersection with the surface, and then, we use the prior color to determine whether this intersection is visible from another view. If it is visible, our multi-view constraint is triggered to make this intersection participate in the rendering along the two rays for better surface inference. For the textureless areas where the multi-view consistency constraint does not work well, we further introduce a depth consistency loss with confidence weights to improve the completeness and smoothness of the surface. Our method does not require additional datasets to learn priors or suffer from generalization issues. Our experimental results outperform the state-of-the-art methods under widely used benchmarks. Our contributions are listed below.
\begin{itemize}
  \item We propose NeRFPrior to reconstruct accurate and smooth scene surfaces by exploiting NeRF as a prior. Such prior is learned by merely overfitting the scene to be reconstructed, without requiring any additional large-scale datasets.
  \item We introduce a novel strategy to impose a multi-view consistency constraint using our proposed NeRFPrior, which reveals more accurate surfaces.
  \item We propose a novel depth consistency loss with confidence weights to improve the smoothness and completeness of reconstructed surfaces for textureless areas in the real-world scenes.
\end{itemize}

\section{Related Work}

\subsection{Multi-view Reconstruction}

Multi-view reconstruction aims at reconstructing 3D surfaces from a given set of uncalibrated multi-view images. Traditional multi-view reconstruction pipeline is split into two stages: the structure-from-motion (SFM)~\cite{schonberger2016structure} and the multi-view-stereo (MVS)~\cite{goesele2006multi}. MVSNet~\cite{yao2018mvsnet} is the first to introduce the learning-based idea into traditional MVS methods. Many following studies improve MVSNet from different aspects, such as training speed~\cite{weilharter2021highres}, memory consumption~\cite{gu2020cascade} and network structure~\cite{ding2022transmvsnet}. 

\subsection{Neural Implicit Reconstruction}

Existing neural implicit reconstruction methods mainly include two categories: reconstruction from point clouds~\cite{ma2021neuralpull, zhou2023uni3d, li2024shsnet} and multi-view images~\cite{Ma2025See3D, wu2025sparis, huang2025fatesgs}. The former typically incorporates various global~\cite{ma2023noise2noise, zhou2024fastnoise2noise, chen2024neuraltps} or local priors~\cite{ma2022pcp,li2024learning, ma2022onsurfacepriors}, along with additional constraints~\cite{chen2024finetuning, zhou2024udiff} or gradients~\cite{noda2024multipull,noda2025bijective,li2024implicit}. However, the optimization relies on ground truth point clouds~\cite{zhou2024capudf,liu2019point2sequence,wen2022pmpnet++,xiang2022snowflake, zhou2023differentiable, li2023neaf}, which are often difficult to acquire. Recently, NeRF~\cite{mildenhall2020nerf} has achieved impressive results in novel view synthesis. Following studies develop the potential of NeRF in various aspects, such as generation~\cite{zhong2023touching, metzer2023latent}, relighting~\cite{yang2023complementary}, human~\cite{chen2022geometry, geng2023learning} and so on. 
Many strategies have been applied to improve the generalization ability~\cite{liu2022neuray, zhang2023fast}, such as integrated positional encoding~\cite{barron2021mipnerf}, voxelization~\cite{chen2022tensorf, sun2022direct} and patch loss~\cite{fu2022geoneus, jiang2025sensing}. 

Recent works~\cite{oechsle2021unisurf, wang2021neus} investigate learning neural implicit fields from multi-view images by differentiable ray marching. More recently, many methods focus on variant kinds of priors to improve the reconstruction quality, for example, depth prior from MVS~\cite{deng2022dsnerf,huang2024neusurf}, ground truth depth~\cite{azinovic2022neuralrgbd, zhang2024learning}, estimated normals from pre-trained models~\cite{yin2024ray, wang2022neuris} and pre-trained semantic segmentation~\cite{zhou2024manhattanpami, zhang2025monoinstance}. Latest neural representation, 3D Gaussian~\cite{huang2025fatesgs,han2024binocular,zhou2024diffgs}, enables SDF inference through splatting~\cite{zhang2024gspull,huang20242dgs,guedon2024sugar,li2025gaussianudf}. However, they struggle to produce plausible surfaces because the geometry and color in 3D is not continuous with 3D Gaussians.

Although the above-mentioned priors can improve the reconstruction quality, there still exists various shortcomings. Data-driven based priors do not generalize well to different kinds of datasets, while overfitting priors can not provide photometric information for the network. To address the above problems, we propose NeRFPrior, which introduces a neural radiance field as a prior to learn implicit functions to reconstruct accurate surfaces without requiring any additional information from large-scale datasets. 

\section{Method}

\begin{figure*}[htbp]
  \centering
  \includegraphics[width=\linewidth]{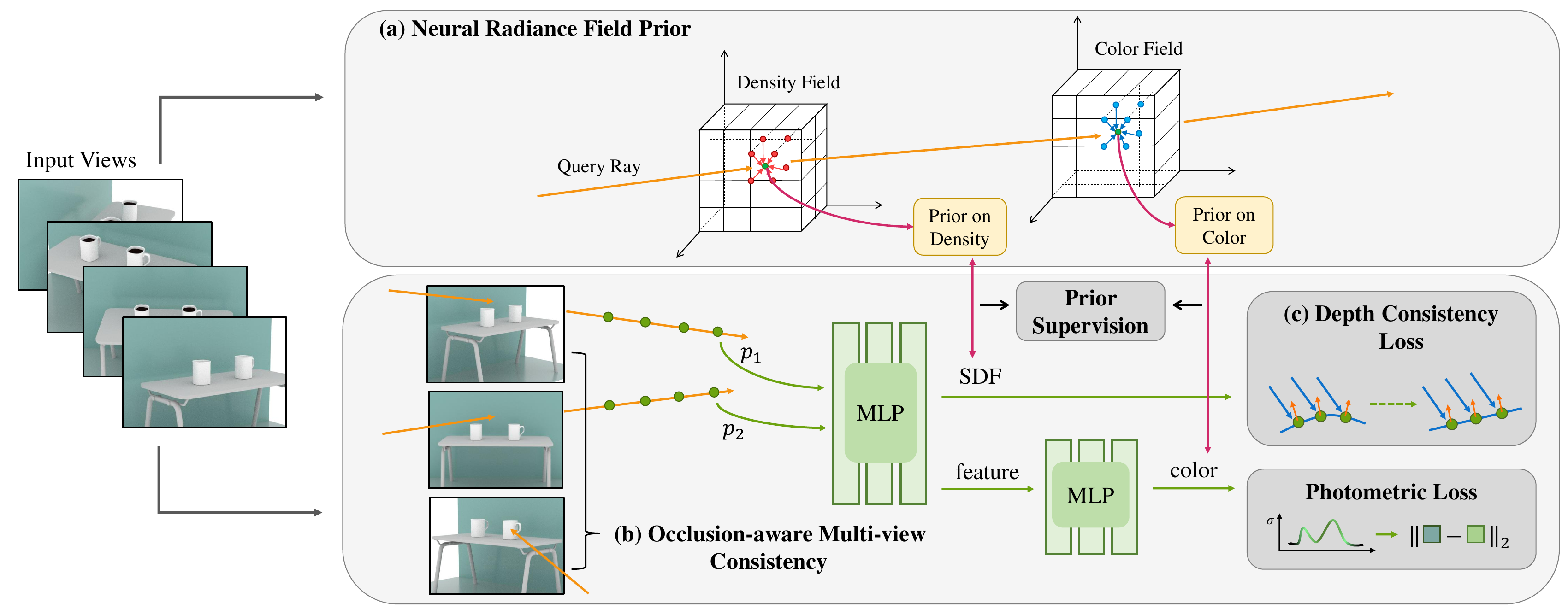}
  \vspace{-0.5cm}
  \caption{An overview of our NeRFPrior method. Given multi-view images of a scene as input, we first train a grid-based NeRF to obtain the density field and color field as priors. We then learn a signed distance function by imposing a multi-view consistency constraint using volume rendering. For each sampled point on the ray, we query the prior density and prior color as additional supervision of the predicted density and color, respectively. To improve the smoothness and completeness of textureless areas in the scene, we propose a depth consistency loss, which forces surface points in the same textureless plane to have similar depths.}
  \label{fig:overview}
  \vspace{-0.3cm}
\end{figure*}
Given a set of posed images captured from a scene, we aim to learn neural implicit functions to reconstruct the scene without requiring any additional information from other datasets. We represent the geometry in the scene as a signed distance field and then extract the mesh using marching cubes algorithm. In this section, we first discuss the insight of adopting neural radiance field as a prior. Then we introduce multi-view consistency constraint and the depth consistency loss with confidence weights as two of our contributions to improve the reconstruction quality. An overview of our framework is provided in Fig.~\ref{fig:overview}.

\begin{figure}[htbp]
  \centering
  \includegraphics[width=\linewidth]{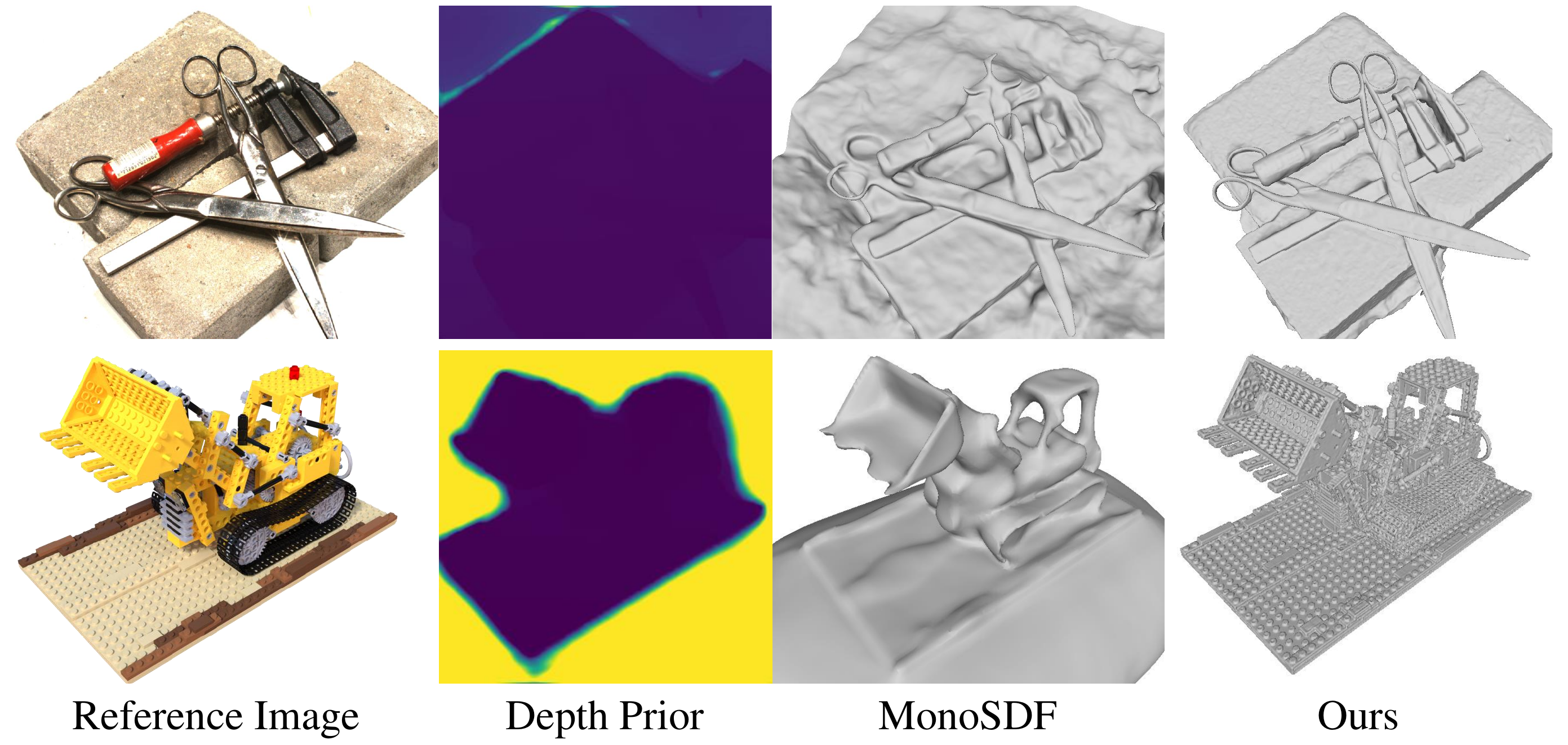}
  \vspace{-0.5cm}
  \caption{Comparison on object-surrounding scenes between MonoSDF and ours. The performance of MonoSDF drastically degenerates because the depth prior cannot generalize well to different kinds of datasets.}
  \label{fig:mono-failure}
  \vspace{-0.3cm}
\end{figure}

\subsection{Neural Radiance Field Prior}

NeRF~\cite{mildenhall2020nerf} models a static scene using a continuous 5D function which takes a 3D coordinate and a corresponding viewing direction as input and outputs per-point density $\sigma$ and color $\mathbf{c}$. Specifically, let $\mathbf{x}_i$ denotes the $i$-th sampled point along the ray $\mathbf{r}$, and $\mathbf{d}$ denotes the viewing direction. The predicted ray color $\hat{C}(\mathbf{r})$ is obtained by volume rendering below:
\begin{equation}
  \begin{gathered}
    \label{eq:volumerendering}
      \hat{C}(\mathbf{r}) = \sum_{i=1}^N T_i(1-\exp(-\sigma_\theta(\mathbf{x}_i) \delta_i)) \mathbf{c}_{\phi}(\mathbf{x}_i, \mathbf{d}) \\ 
      T_i = \exp(-\sum_{k=1}^{i-1} \sigma_\theta(\mathbf{x}_k) \delta_k),
  \end{gathered}
\end{equation}
\noindent where $\delta_i$ and $T_i$ represent the sampling interval and the accumulated transmittance of the ray $\mathbf{r}$ at $i$-th sampled point, respectively. $\theta$ and $\phi$ are the parameters of the density and color networks, respectively. 

Recently, there has been a number of studies combining NeRF framework and implicit functions to reconstruct 3D surfaces. However, the advanced NeRF techniques~\cite{chen2022tensorf, sun2022direct, muller2022instant, fridovich2022plenoxels} inspire us that NeRF itself can serve as a prior for surface reconstruction. Compared to NeRF-based surface reconstruction methods~\cite{wang2021neus,oechsle2021unisurf,zhou2024manhattanpami}, we have the ability to explicitly use geometry and color information from the field for visibility check and imposing multi-view depth consistency constraints. This design has two main advantages. Firstly, our NeRF prior is able to provide color cues for optimization, which is missing in other methods combining priors~\cite{yu2022monosdf, fu2022geoneus}.

Secondly, our prior is easily accessed compared to existing prior acquisition methods. Data-driven priors such as depth and normal priors~\cite{yu2022monosdf, wang2022neuris, wang2022neuralroom}, need days of pre-training on large-scale datasets. Additionally, data-driven priors do not generalize well to different kinds of scenes, as shown in Fig.~\ref{fig:mono-failure}. The prior of MonoSDF is pretrained on indoor scene datasets, so the quality of prior degenerates while generalizing to object-surrounding datasets.
On the other hand, overfitting priors such as sparse depth and sparse point cloud from COLMAP algorithm~\cite{fu2022geoneus, wang2022neuralroom, guo2022neural}, are sparse and incontinuous that most pixels or points cannot be supervised. And it lacks the supervision of color. Thanks for the advance in NeRF training acceleration, we can optimize a grid-based NeRF, which can be trained in minutes. Additionally, the grid-based structure has advantages in perceiving high-frequency surface details, which is beneficial to our accurate reconstruction.

As shown in Fig.~\ref{fig:overview} (a), to obtain the neural radiance field prior from multi-view images, we firstly construct a pair of density grid $F_\sigma \in \mathbb{R}^{[N_1, N_2, N_3, 1]}$ and color feature grid $F_\mathbf{c} \in \mathbb{R}^{[N_1, N_2, N_3, d]}$, where $N_1, N_2, N_3$ are the resolutions of the feature grids, and $d$ is the feature length of color grid. For a 3D point $\mathbf{x}$ sampled along the rendering ray with viewing direction $\mathbf{d}$, the density and color are interpolated from the feature grids of the trained NeRF, as denoted by
\begin{equation}
    \begin{gathered}
        \sigma_{prior}(\mathbf{x}) = \operatorname{act} (\operatorname{interp} (F_\sigma, \mathbf{x})) \\
        \mathbf{c}_{prior}(\mathbf{x}, \mathbf{d}) = \operatorname{act} (\operatorname{MLP}(\operatorname{interp} (F_\mathbf{c}, \mathbf{x}), \mathbf{d})),
    \end{gathered}
\end{equation}
\noindent where the operation $\operatorname{act}$ represents activation function and $\operatorname{interp}$ represents trilinear interpolation, respectively. For color prediction, we use an additional shallow MLP to take viewing direction into consider. The network is trained using volume rendering and then frozen as our NeRF prior.

Following~\cite{wang2021neus}, we further integrate the signed distance field into neural surface reconstruction by learning SDF to represent density in volume rendering:
\begin{equation}
\label{eqn:neus-density2sdf}
  \sigma(\textbf{x})=\max \left( \frac{-\Phi'(f_s(\textbf{x}))}{\Phi(f_s(\textbf{x}))}, 0 \right),
\end{equation}
\noindent where $\mathbf{x}$ represents the sampled point along the ray. $\Phi$ and $f_s$ represent sigmoid function and SDF network, respectively. To combine the prior field and the signed distance field together, we query the density and color of each sampled point from the prior fields and use them as supervision of the predicted density and color from neural implicit network:
\begin{equation}
    \label{eq: priorloss}
    \begin{gathered}
        \mathcal{L}_{\sigma}=\Vert \hat{\sigma}(\textbf{x})-\sigma_{prior}(\textbf{x}) \Vert_1 \\
        \mathcal{L}_{c}=\Vert \hat{\textbf{c}}(\textbf{x}, \textbf{d})-\textbf{c}_{prior}(\textbf{x}, \textbf{d}) \Vert_1 .
    \end{gathered}
\end{equation}
We notice that the prior density field is usually noisy, which may mislead the neural implicit network. Therefore, we use a threshold to filter out the fuzzy density value and apply supervision only if the density value is convincing. The filtering strategy will be discussed in the supplementary in detail. Benefiting from the NeRF prior, we are able to learn the signed distance field to reconstruct accurate 3D geometry details at a fast speed.

\subsection{Multi-view Consistency Constraint}
\begin{figure}[t]
  \centering
  \includegraphics[width=\linewidth]{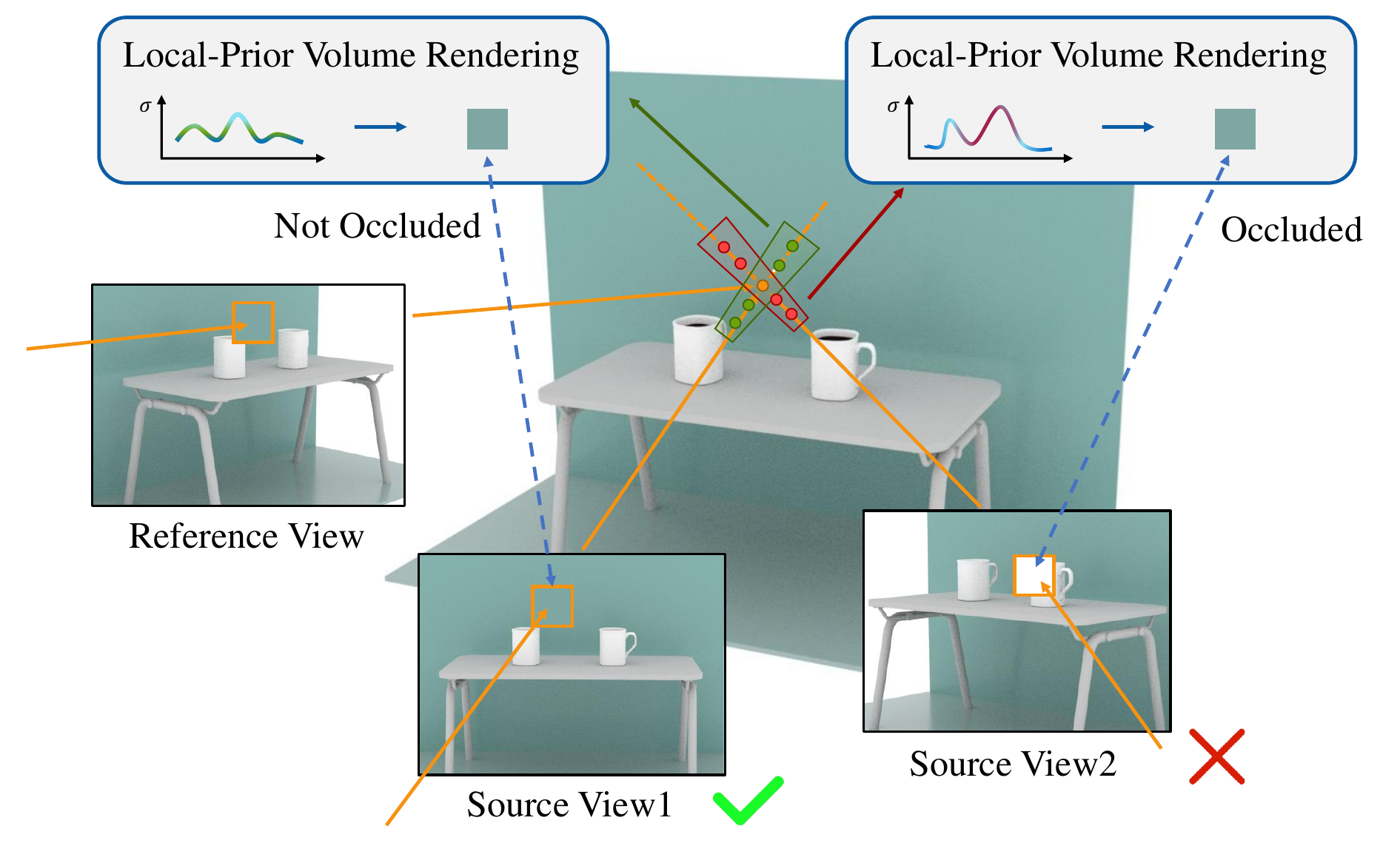}
  \vspace{-0.6cm}
  \caption{An illustration of our multi-view consistency constraint. To judge the visibility of the intersection, we conduct a local-prior volume rendering around the intersection and compare the rendering color with the projection color. The ray from source view is participated in training only if the intersection is visible along this ray.}
  \vspace{-0.2cm}
  \label{fig:multiview}
\end{figure}
Multi-view consistency is a key intuition for geometry extraction because the photometric consistency information existed in the multi-view images is a powerful prompt to help revealing the surface. To reconstruct accurate 3D surfaces, we explicitly impose a multi-view consistency constraint on each ray for its SDF inference. Specifically, for an emitted ray $\mathbf{r}_m$ from a reference view $I_m$, we firstly apply root finding~\cite{oechsle2021unisurf} to locate the intersection point $\mathbf{p}^*$ where the ray hits the surface. Then we select several nearby images as source views. For each source view, we emit an additional ray from the camera viewpoint to the intersection $\mathbf{p}^*$. The ray from reference view and the rays from source views are gathered and fed into volume rendering in parallel. An intuition of this idea is that the network is enabled to inference the zero-level-set of the intersection from the photometric difference of multi-view images, as shown in Fig.~\ref{fig:overview} (b) and detailed in Fig.~\ref{fig:multiview}.
\begin{figure}[t]
  \centering
  \includegraphics[width=\linewidth]{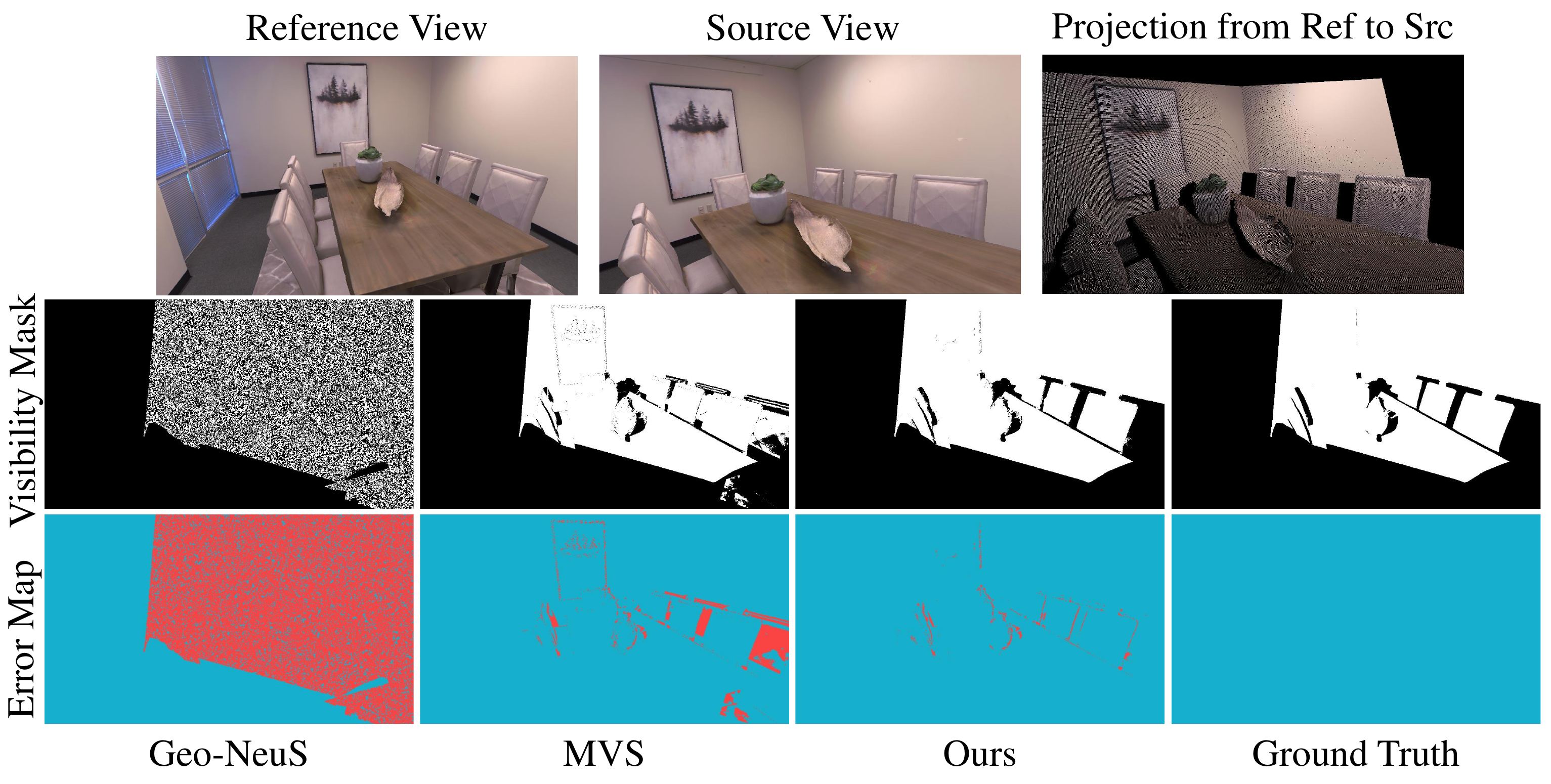}
  \vspace{-0.6cm}
  \caption{A comparison on the accuracy of visibility check. The first row shows the ground truth result of projecting pixels from reference view to source view. The second row shows the visibility mask, indicating which points in the reference view are visible after projection. The third row is the error map of visibility check. }
  \vspace{-0.2cm}
  \label{fig:multiview-visibility check}
\end{figure}
While emitting multi-view rays towards an intersection, some rays may be blocked by some objects in front of the intersection. To resolve this issue, we use our prior field to conduct a local-prior volume rendering for visibility check. Specifically, to determine the visibility of intersection $\mathbf{p}^*$ from source view  $I_{s}$ with viewing direction $\mathbf{r}_{s}$, we sample $M$ points in a small interval $[d^*_{s}-\Delta, d^*_{s}+\Delta]$ centered at $\mathbf{p}^*$ along $\mathbf{r}_{s}$, where $d^*_{s}$ is the distance between $\mathbf{p}^*$ and the viewpoint of $I_{s}$. Next we apply volume rendering on the sampled points using the queried prior density and prior color: 
\begin{equation}
  \begin{gathered} 
      \mathbf{c}^*_{s} = \sum_{k=1}^M T_k(1-\exp(-\sigma_{prior}(\mathbf{x}_k) \delta)) \mathbf{c}_{prior}(\mathbf{x}_k, \mathbf{d}(\mathbf{r}_s)), \\ 
      T_k = \exp(-\sum_{q=1}^{k-1} \sigma_{prior}(\mathbf{x}_q) \delta),
  \end{gathered}
\end{equation}
\noindent where $\mathbf{d}(\mathbf{r}_s)$ represents the viewing direction of $\mathbf{r}_s$. In practice, we typically set $\Delta=0.1$, $M=64$ and $\delta=0.003$. The rendered color $\mathbf{c}^*_{s}$ is compared with the pixel color $\mathbf{c}^{proj}_{s}$, which is the projection of $\mathbf{p}^*$ on the source view $I_{s}$. If the two colors differ a lot, we consider that ${\mathbf{p}^*}$ is invisible from $I_{s}$, otherwise visible:
\begin{equation}
\label{eq: visibility-check}
\mathbf{p}^*=\left\{
    \begin{array}{cl}
        \text{visible} &  |\mathbf{c}^*_{s}-\mathbf{c}^{proj}_{s}|<t_0 \\
        \text{invisible}  &  |\mathbf{c}^*_{s}-\mathbf{c}^{proj}_{s}|\ge t_0 \\
    \end{array}
    \right.
\end{equation}
If $\mathbf{p}^*$ is visible, we then emit the ray $\mathbf{r}_{s}$ for volume rendering together with the ray $\mathbf{r}_m$ from the reference view.

Our visibility check is more robust than traditional MVS methods which directly match the projection color on two views, since the color of projections is significantly biased on illumination. Our NeRFPrior resolves this issue by predicting view-dependent color. Although the standard volume rendering needs sampling in a fairly long interval, we observe that due to the pulse characteristics of density, only a small interval is enough for volume rendering to get accurate color in the pretrained NeRF. Fig.~\ref{fig:multiview-visibility check} provides an example. Comparing to Geo-NeuS~\cite{fu2022geoneus} which uses patched normalization cross correlation (NCC) to judge visibility and MVS~\cite{schonberger2016structure} which depends on projection color to judge visibility, our method achieves significantly more accurate results.

\subsection{Depth Consistency Loss}
\begin{figure}[htbp]
  \centering
  \includegraphics[width=\linewidth]{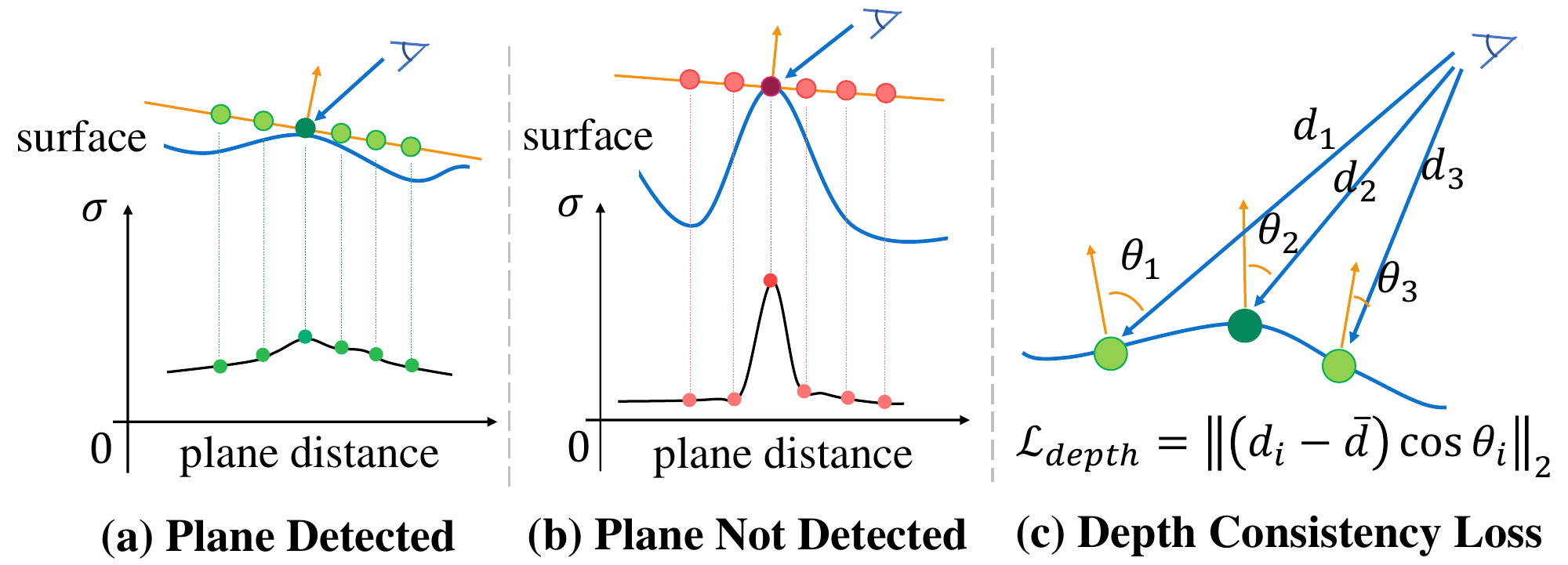}
  \vspace{-0.5cm}
  \caption{An illustration of our depth consistency loss. We calculate the density variance of the intersection and its neighboring points on the tangent plane. If (a) the variance is small, we constrain these points to maintain the same depth on normal directions as in (c). Otherwise, (b) we do not impose depth constraints.}
  \vspace{-0.2cm}
  \label{fig:depthloss}
\end{figure}
It is hard for neural implicit functions to infer accurate surfaces in textureless areas in indoor scenes such as walls and floors, due to the lack of distinctive color information. We further propose a depth consistency loss with confidence weights to improve the smoothness and completeness in textureless areas. We observe that continuous textureless areas usually have consistent or continuously varying colors, and are usually composed of planes~\cite{wang2022neuralroom}. Hence, we use density distribution as a clue to determine whether the neighboring area of an intersection is a plane, and then add depth consistency constraints if it is the case, as shown in Fig.~\ref{fig:overview} (c) and detailed in Fig.~\ref{fig:depthloss}.

In order to impose depth consistency constraints on surface points, two prerequisites are needed: (i) the intersection and its neighboring points have similar colors on the projection view, (ii) the intersection and its neighboring points are nearly on a plane. For (i), we calculate the color variance of each pixel and its neighboring pixels on the input views. For (ii), we calculate density variance of the intersection and its neighboring points as a confidence to judge whether a surface is a plane. If the density variance and the color variance are both small, we assume that the ray hits a plane. Then we constrain the neighborhood points to maintain the same depth on their normal directions. Otherwise, we do not impose depth constraints. Formally, let $\mathbf{p}^*$ be the intersection between ray $\mathbf{r}$ and the object surface, $\mathbf{c}^{proj}$ be the projection pixel color of $\mathbf{p}^*$ on the source view. The depth loss can be written as following: 
\begin{equation}
    \label{eq: depthloss}
    \mathcal{L}_{depth} = \sum\limits_{\mathbf{r}\in \mathcal{R}} \Vert (\hat{D}(\mathbf{r})-\Bar{D})  \cos\left\langle \mathbf{n}, \mathbf{r} \right\rangle \Vert_2 * \operatorname{sgn}_{c} * \operatorname{sgn}_{\sigma} 
\end{equation}
\begin{equation}
\label{eq: depthloss2}
    \begin{gathered}
        \operatorname{sgn}_{c}=\left\{
        \begin{array}{cl}
        1 & \varr(\mathbf{c}^{proj})<t_1 \\
        0 & \varr(\mathbf{c}^{proj})\ge t_1 \\
        \end{array}
        \right. \\
        \operatorname{sgn}_{\sigma}=\left\{
        \begin{array}{cl}
        1 & \varr(\sigma(\mathbf{p}^*))<t_2 \\
        0 & \varr(\sigma(\mathbf{p}^*))\ge t_2 \\
        \end{array}
        \right.
    \end{gathered}
\end{equation}
\noindent where $\hat{D}(\mathbf{r})$ is the rendered depth of ray $\mathbf{r}$ and $\Bar{D}$ is the mean depth in a batch of rays $\mathcal{R}$, which are emitted from some neighboring pixels. $\mathbf{n}$ is the rendered normal vector of ray $\mathbf{r}$, and $\varr$ represents the variation. In a word, only when the intersection is on a plane and it is in the textureless areas of the image, we constrain the depth of the intersection to keep similar with the depth of its neighboring intersections.

\subsection{Loss Function}

We render the color of each ray using Eq.~(\ref{eq:volumerendering}) and measure the error between rendered color and ground truth pixel color:
\begin{equation}
    \mathcal{L}_{rgb} = \sum_{\mathbf{r}\in \mathcal{R}}\Vert \hat{C}(\mathbf{r}) - C(\mathbf{r}) \Vert_1,
\end{equation}
\noindent where $\mathcal{R}$ denotes all of the rays in a training batch. Following~\cite{wang2021neus}, we add an Eikonal term on the sampled points to regularize the SDF field by 
\begin{equation}
\label{eq: normalloss}
    \mathcal{L}_{reg} = \frac{1}{N} \sum_{i}\Vert \nabla f_s(\mathbf{p}_i)-1 \Vert_2,
\end{equation}
\noindent where $\mathbf{p}_i$ is the sampled point on the ray and $N$ is the number of sampled points.

With our additional prior field supervision (Eq.~\ref{eq: priorloss}) and depth loss (Eq.~\ref{eq: depthloss}), the overall loss function can be written as
\begin{equation}
\label{eq: totalloss}
    \mathcal{L} = \mathcal{L}_{rgb} + \lambda_1\mathcal{L}_\sigma+\lambda_2\mathcal{L}_{c} + \lambda_3 \mathcal{L}_{reg} + \lambda_4 \mathcal{L}_{depth}.
\end{equation}
%


%
\begin{figure*}[t]
  \centering
  \includegraphics[width=\linewidth]{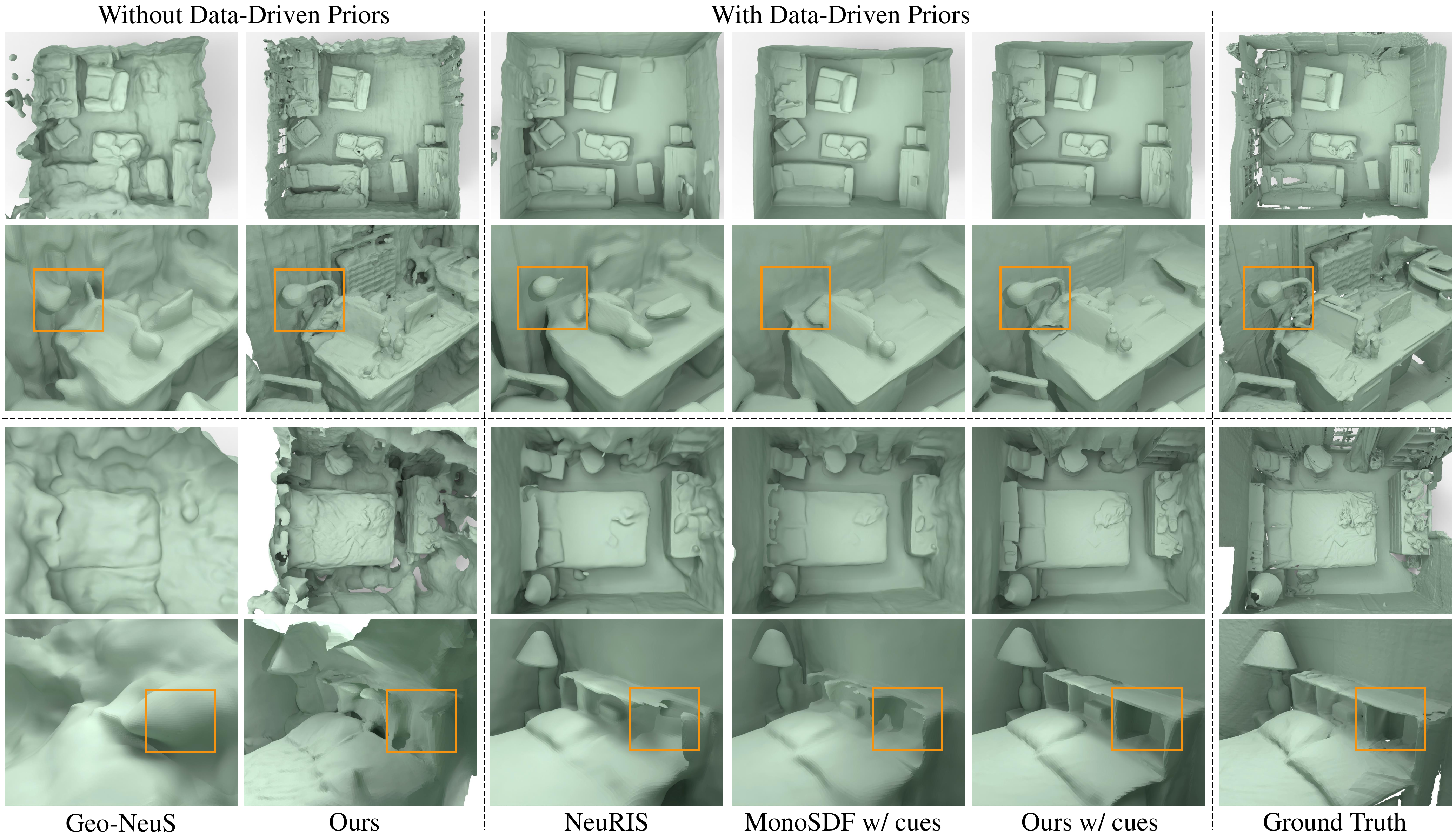}
  \vspace{-0.6cm}
  \caption{Visualization comparison on ScanNet Dataset.}
  \vspace{-0.2cm}
  \label{fig:scannet}
\end{figure*}
\begin{figure*}[t]
  \centering
  \includegraphics[width=\linewidth]{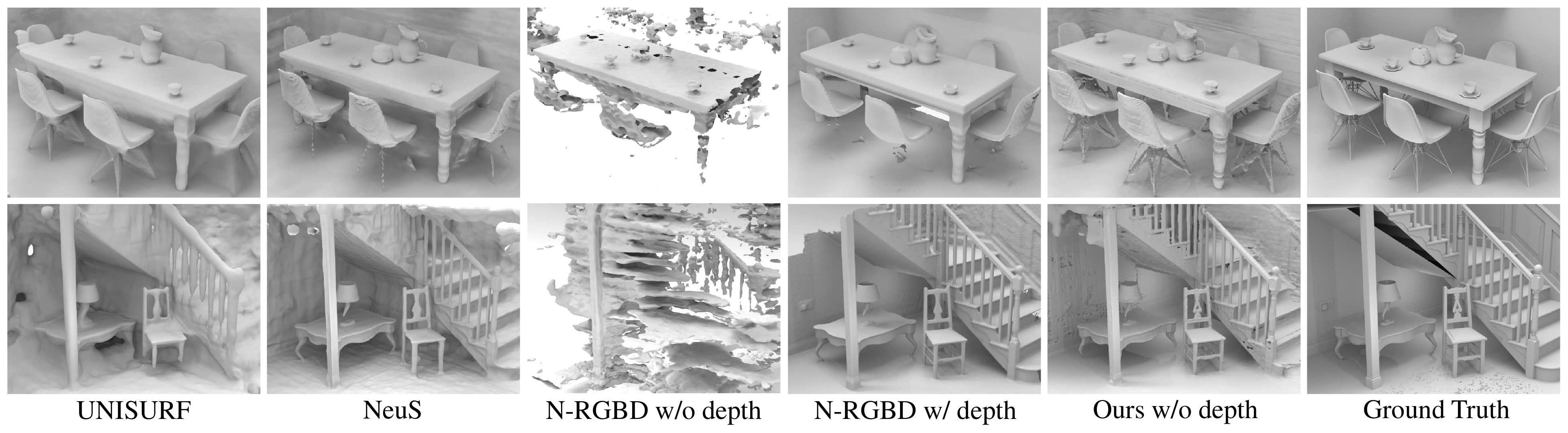}
  \vspace{-0.6cm}
  \caption{Visualization comparison on BlendSwap Dataset.}
  \vspace{-0.3cm}
  \label{fig:blendswap}
\end{figure*}
\begin{figure*}[t]
  \centering
  \includegraphics[width=\linewidth]{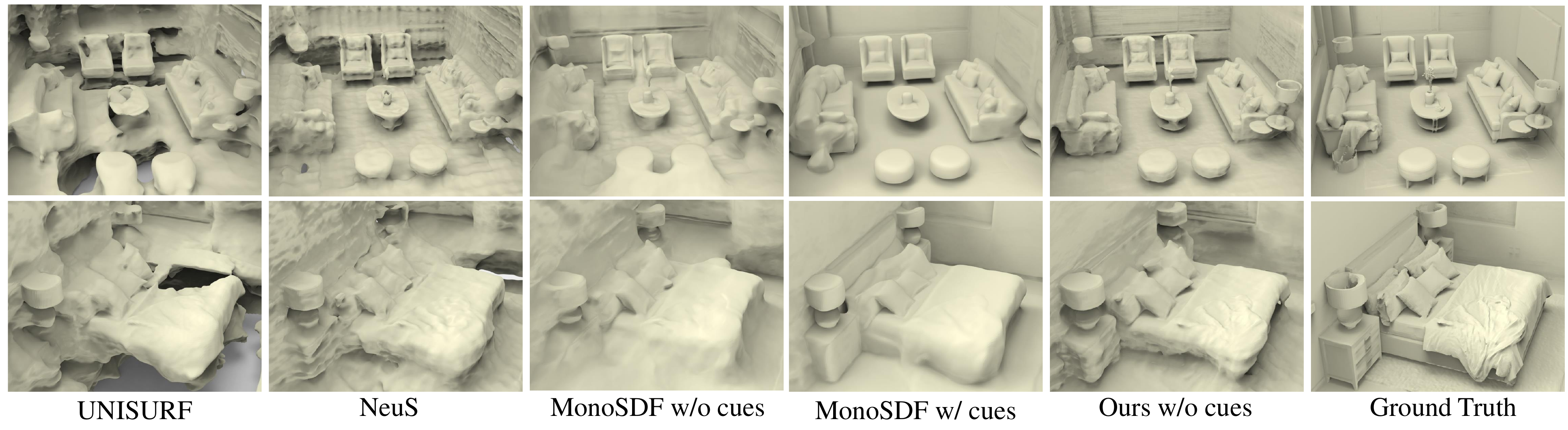}
  \vspace{-0.6cm}
  \caption{Visualization comparison on Replica Dataset.}
  \vspace{-0.3cm}
  \label{fig:replica}
\end{figure*}

\section{Experiments}

\subsection{Implementation Details}
To train a neural radiance field as our NeRF prior, we adopt the grid-based architecture of TensoRF~\cite{chen2022tensorf}. We train the prior NeRF for each scene in 30k iterations, which takes about 30 minutes per scene. For our implicit surface function, we adopt the architecture of NeuS~\cite{wang2021neus}, where the signed distance function and color function are modeled by an MLP with 8 and 6 hidden layers, respectively. We train our implicit surface function for 200k iterations in total. The multi-view consistency constraint is applied after 100k iterations and the depth consistency loss is applied after 150k iterations. We adopt such strategy based on the observation that the multi-view consistency and depth loss may mislead the network at the early training stage when the surface is noisy and ambiguous. We set $t_0=0.02$ in Eq.~(\ref{eq: visibility-check}), $t_1=0.04$ and $t_2=0.1$ in Eq.~(\ref{eq: depthloss2}), $\lambda_1=\lambda_2=0.1$ and decreases exponentially to 0, $\lambda_3=0.05$ and $\lambda_4=0.5$ in Eq.~(\ref{eq: totalloss}). The choice of hyperparameters and thresholds will be discussed in supplementary in details. All the experiments are conducted on a single NVIDIA RTX 3090Ti GPU.

\subsection{Experimental Settings}

\noindent \textbf{Datasets.} \ We evaluate our method quantitatively and qualitatively on real-captured dataset ScanNet~\cite{dai2017scannet}. Following previous works~\cite{yu2022monosdf}, we use 4 scenes from ScanNet for our evaluation. We also evaluate our method under two synthetic scene datasets, including BlendSwap~\cite{azinovic2022neuralrgbd} and Replica~\cite{straub2019replica}, each of which contains 8 indoor scenes.

\noindent \textbf{Baselines.}  We compare our method with the following state-of-the-art methods: (1) Classic MVS method: \textbf{COLMAP}~\cite{schonberger2016structure}. (2) Neural raidance field methods without data-driven priors: \textbf{NeRF}~\cite{mildenhall2020nerf}, \textbf{UNISURF}~\cite{oechsle2021unisurf}, \textbf{NeuS}~\cite{wang2021neus}, \textbf{Geo-NeuS}~\cite{fu2022geoneus}, \textbf{PermutoSDF}~\cite{rosu2023permutosdf}, \textbf{NeuralAngelo}~\cite{li2023neuralangelo}. (3) Neural implicit reconstruction methods with data-driven priors: \textbf{Neural RGB-D}~\cite{azinovic2022neuralrgbd}, \textbf{Manhattan-SDF}~\cite{guo2022neural}, \textbf{NeuRIS}~\cite{wang2022neuris}, \textbf{MonoSDF}~\cite{yu2022monosdf}, \textbf{GO-Surf}~\cite{wang2022gosurf}. 

\noindent \textbf{Evaluation Metrics.} \ For ScanNet dataset, following~\cite{yu2022monosdf}, we adopt Accuracy, Completeness, Precision, Recall and F1-score as evaluation metrics. For synthetic dataset, following~\cite{azinovic2022neuralrgbd}, we adopt Chamfer Distance (CD), Normal Consistency (NC), Precision, Recall and F1-score as evaluation metrics. Please refer to the supplementary for more details on these metrics.
\begin{table}[t]
  \centering
  \caption{Evaluation results on ScanNet dataset. MonoSDF$^*$ represents MonoSDF with its monocular depth and normal cues.}
  \vspace{-0.2cm}
  \label{tab:scannet compare}
  \resizebox{\linewidth}{!}{\begin{tabular}{l|ccccc}
    \toprule
        Methods & Acc $\downarrow$ & Comp $\downarrow$ & Prec $\uparrow$ & Recall $\uparrow$ & F1 $\uparrow$ \\
    \midrule
         NeRF\cite{mildenhall2020nerf}  & 0.735 & 0.177 & 0.131 & 0.291 & 0.176   \\
         NeuS\cite{wang2021neus}        & 0.179 & 0.208 & 0.313 & 0.275 & 0.291   \\
         Geo-Neus\cite{fu2022geoneus}       & 0.236 & 0.206 & 0.282 & 0.313 & 0.291   \\
         MonoSDF\cite{yu2022monosdf}    & 0.214 & 0.180 & 0.297 & 0.325 & 0.310   \\
         PermutoSDF\cite{rosu2023permutosdf} & 0.143 & 0.219 & \textbf{0.448} & 0.209 & 0.285 \\
         NeuralAngelo\cite{li2023neuralangelo} & 0.245 & 0.272 & 0.274 & 0.311 & 0.292 \\
         Ours                           & \textbf{0.133} & \textbf{0.120} & 0.439 & \textbf{0.429} & \textbf{0.433}   \\
   \cmidrule{1-6} 
         Manhattan\cite{guo2022neural}  & 0.072 & 0.068 & 0.621 & 0.586 & 0.602   \\
         NeuRIS\cite{wang2022neuris}    & 0.054 & 0.052 & 0.729 & 0.684 & 0.705   \\
         MonoSDF$^*$\cite{yu2022monosdf}   & 0.042 & 0.049 & 0.760 & 0.707 & 0.732   \\
         Ours\ (+monocular cues)                          & \textbf{0.037} & \textbf{0.042} & \textbf{0.799} & \textbf{0.766} & \textbf{0.782}  \\
     \cmidrule{1-6} 
         Go-Surf\cite{wang2022gosurf}  & 0.048 & 0.021 & 0.880 & 0.894 & 0.887   \\
         Ours\ (+depth)                       & \textbf{0.027} & \textbf{0.020} & \textbf{0.931} & \textbf{0.928} & \textbf{0.930}  \\
  \bottomrule
\end{tabular}}
\vspace{-0.3cm}
\end{table}

\begin{table}[t]
  \centering
  \caption{Evaluation results on BlendSwap dataset. Results are averaged among the 8 scenes.}
  \vspace{-0.2cm}
  \label{tab:blendswap compare}
  \resizebox{\linewidth}{!}{\begin{tabular}{l|ccccc}
    \toprule
        Methods & CD $\downarrow$ & NC $\uparrow$ & Prec $\uparrow$ & Recall $\uparrow$ & F1 $\uparrow$ \\
    \midrule
        COLMAP\cite{schonberger2016structure} & 0.420 & 0.556 & 0.429 & 0.353 & 0.387   \\
        UNISURF\cite{oechsle2021unisurf}      & 0.213 & 0.710 & 0.610 & 0.413 & 0.484   \\
        NeuS\cite{wang2021neus}               & 0.180 & 0.731 & 0.526 & 0.454 & 0.483   \\
        N-RGBD\cite{azinovic2022neuralrgbd}       & 0.380 & 0.423 & 0.266 & 0.219 & 0.292   \\
    \cmidrule{1-6}
        Ours    & \textbf{0.088} & \textbf{0.813} & \textbf{0.651} & \textbf{0.594} & \textbf{0.621}  \\
  \bottomrule
\end{tabular}}
\vspace{-0.3cm}
\end{table}

\begin{table}[t]
  \centering
  \caption{Evaluation results on Replica dataset. Results are averaged among the 8 scenes.}
  \vspace{-0.2cm}
  \label{tab:replica compare}
  \resizebox{\linewidth}{!}{\begin{tabular}{l|ccccc}
    \toprule
        Methods & CD $\downarrow$ & NC $\uparrow$ & Prec $\uparrow$ & Recall $\uparrow$ & F1 $\uparrow$ \\
    \midrule
        COLMAP\cite{schonberger2016structure}  & 0.232 & 0.468 & 0.455 & 0.408 & 0.430   \\
        UNISURF\cite{oechsle2021unisurf}       & 0.110 & 0.769 & 0.566 & 0.449 & 0.496   \\
        NeuS\cite{wang2021neus}                & 0.066 & 0.883 & 0.709 & 0.626 & 0.665   \\
        MonoSDF\cite{yu2022monosdf}            & 0.075 & 0.867 & 0.657 & 0.609 & 0.632   \\
    \cmidrule{1-6}
        Ours     & \textbf{0.038} & \textbf{0.912} & \textbf{0.833} & \textbf{0.795} & \textbf{0.813}  \\
  \bottomrule
\end{tabular}}
\vspace{-0.3cm}
\end{table}
\begin{table}[t]
  \centering
  \caption{Comparison of the total time of training pipeline.}
  \vspace{-0.2cm}
  \label{tab:time compare}
  \resizebox{\linewidth}{!}{\begin{tabular}{lccc}
    \toprule
      Methods & Getting Priors & Training & Total \\
    \midrule
        COLMAP\cite{schonberger2016structure}  & 10.7h & - & 8.7h  \\
        NeuS\cite{wang2021neus}                & - & 7.2h & 7.2h    \\
        Neural RGB-D\cite{azinovic2022neuralrgbd}  & - & 10.3h & 10.3h    \\
        Geo-NeuS\cite{fu2022geoneus}                   & 1.5h & 7.5h & 9.0h \\
        MonoSDF\cite{yu2022monosdf}            & - & 10.6h & 10.6h    \\
    \cmidrule{1-4}
        Ours          & 37min & 4.2h & \textbf{4.7h}  \\
  \bottomrule
\end{tabular}}
\vspace{-0.3cm}
\end{table}
\subsection{Quantitative and Qualitative Comparison}

\noindent\textbf{Evaluation on ScanNet Dataset.} We report our evaluation on ScanNet dataset in Tab.~\ref{tab:scannet compare} and Fig.~\ref{fig:scannet}. The comparison is split into three parts. The first part is the comparison with the methods that do not use data-driven priors, including NeRF, NeuS, Geo-NeuS, MonoSDF without cues, PermutoSDF. The second part is the comparison with the methods that use data-driven priors, including Manhattan with pretrained segmentation priors, NeuRIS with pretrained normal priors, MonoSDF with estimated depth and normal cues (marked as ``MonoSDF$^*$''), and our results integrated with MonoSDF cues. The third part is the comparison with the methods that use ground truth depth supervision, including Go-Surf and our results with depth supervision. Our method exceeds other baselines without data-driven priors. On the other hand, integrated with monocular cues or ground truth depth supervision, our method also achieves the best performance comparing to other methods with priors. Visual comparisons in Fig.~\ref{fig:scannet} show that our method is able to reconstruct complete and smooth surfaces and captures more scene details, such as the lamp and the bedside cupboard. 

\noindent\textbf{Evaluation on BlendSwap Dataset. }We report our evaluation on BlendSwap dataset in Tab.~\ref{tab:blendswap compare} and Fig.~\ref{fig:blendswap}. 
We compare our method with state-of-the-art methods that do not use data-driven priors, including COLMAP, UNISURF, NeuS and Neural-RGBD without ground truth depth supervision (marked as ``N-RGBD''). The results show our brilliant ability of inferring implicit representations from multi-view images. Additionally, our advantages over our baseline ``NeuS'' highlight the benefits we get from the NeRF prior. Visual comparisons in Fig.~\ref{fig:blendswap} show that our reconstruction does not have artifacts, and contains more details with much higher accuracy than other methods. 

\noindent\textbf{Evaluation on Replica Dataset. }We evaluate our method on Replica dataset, as shown in Tab.~\ref{tab:replica compare} and Fig.~\ref{fig:replica}. We report comparisons with the latest methods, including COLMAP, UNISURF, NeuS and MonoSDF without cues. Qualitative results in Fig.~\ref{fig:replica} further demonstrate the advantages of our method on reconstructing complete, smooth and high fidelity surfaces.

\noindent\textbf{Optimization Time. }We evaluate the total time of training pipeline of different methods, including the time of obtaining priors and the time of training, as reported in Tab.~\ref{tab:time compare}. Benefiting from the advance in NeRF training acceleration~\cite{chen2022tensorf}, we are able to obtain our NeRF prior in half an hour, comparing to COLMAP which takes a long time in dense reconstruction. With the guidance of the NeRF prior, our network is able to converge fast in the early stage of training, which reduces the total training time by about 50\% compared to current neural implicit function methods.
\begin{table}[t]
  \centering
  \caption{Ablation study on each module of our method.}
  \vspace{-0.1cm}
  \label{tab: ablation-compare}
  \resizebox{\linewidth}{!}{\begin{tabular}{ccccc|ccc}
    \toprule
        Base & NeRF prior & Multi-view & Depth loss & Reg term & CD $\downarrow$ & NC $\uparrow$ & F1 $\uparrow$ \\
    \midrule
        \checkmark &            &            &            & \checkmark & 0.083 & 0.832 & 0.619   \\
        \checkmark &            & \checkmark & \checkmark & \checkmark & 0.051 & 0.893 & 0.781   \\
                   & \checkmark &            &            & \checkmark & 0.049 & 0.763 & 0.673   \\
        \checkmark & \checkmark &            &            & \checkmark & 0.050 & 0.887 & 0.744   \\
        \checkmark & \checkmark & \checkmark &            & \checkmark & 0.044 & 0.897 & 0.773   \\
        \checkmark & \checkmark & \checkmark & \checkmark &            & 0.043 & 0.873 & 0.794   \\
        \checkmark & \checkmark & \checkmark & \checkmark & \checkmark & \textbf{0.038} & \textbf{0.912} & \textbf{0.813} \\
  \bottomrule
\end{tabular}}
\vspace{-0.3cm}
\end{table}
\begin{figure}[ht]
  \includegraphics[width=\linewidth]{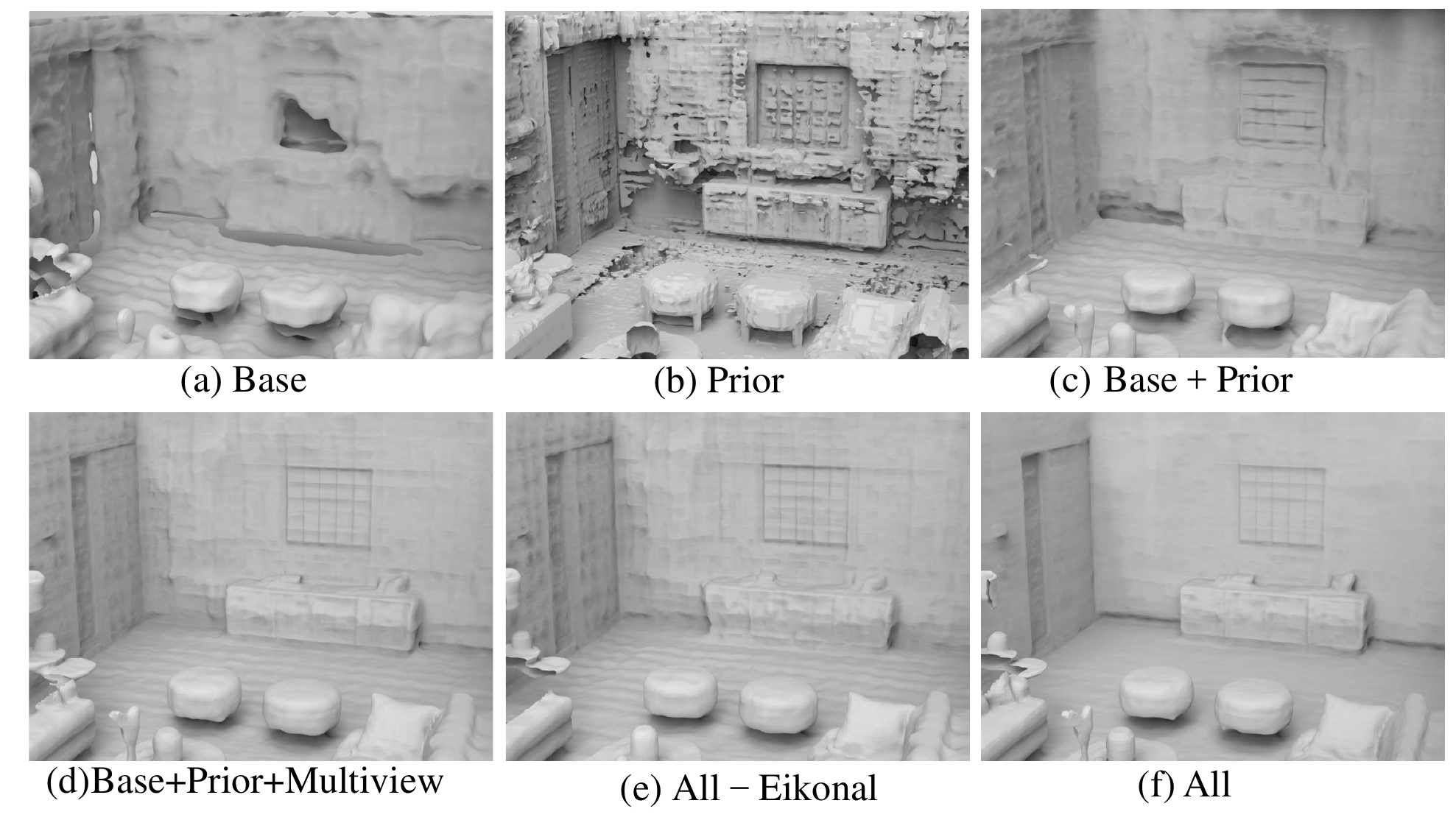}
  \vspace{-0.6cm}
  \caption{Ablation study on each module of our method.}
  \vspace{-0.4cm}
  \label{fig: ablation-allmodules}
\end{figure}
\begin{figure}[t]
  \centering
  \includegraphics[width=\linewidth]{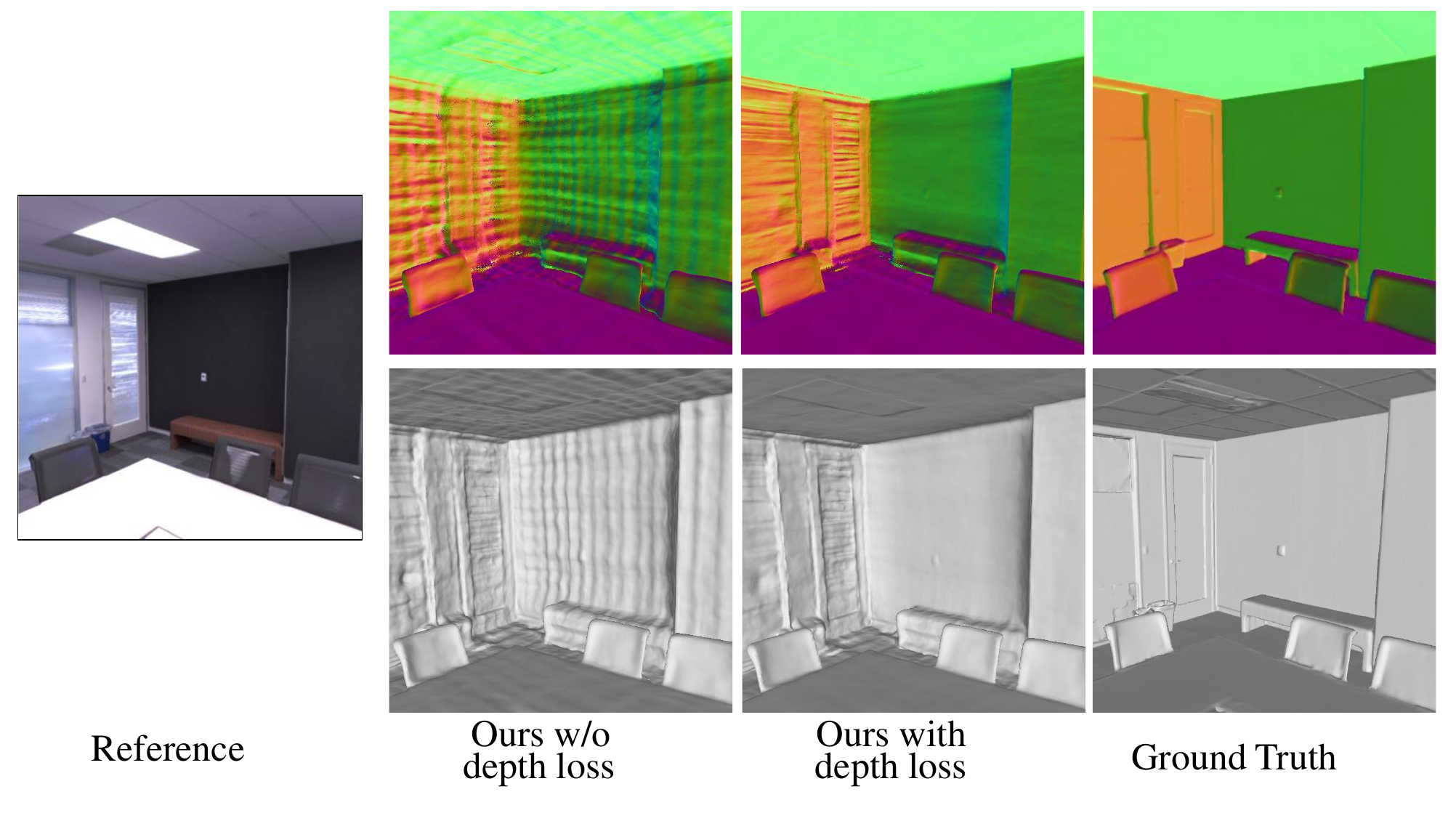}
  \vspace{-0.7cm}
  \caption{A visualization of the ablation on depth consistency loss. The first line is the normal map and the second line is the reconstructed mesh.}
  \vspace{-0.3cm}
  \label{fig: ablation-depthloss-2}
\end{figure}
\subsection{Ablation Study}

To demonstrate the effectiveness of our proposed components, we conduct ablation studies on Replica dataset, as reported in Tab.~\ref{tab: ablation-compare} and Fig.~\ref{fig: ablation-allmodules}. We report our visualization and quantification results on 6 different settings: \textbf{(a)} only the base implicit function network, \textbf{(b)} only the NeRF prior, \textbf{(c)} the base network with our NeRF prior, \textbf{(d)} the base network with NeRF prior and the multi-view consistency constraint, \textbf{(e)} the complete method without eikonal regularization term, \textbf{(f)} our complete method. Our NeRF prior is able to perceive geometric details but shows very poor performance on consistency and smoothness, as shown in Fig.~\ref{fig: ablation-allmodules} (b). With the help of multi-view consistency constraint and depth consistency loss, we can reconstruct high fidelity scene surfaces.

We further conduct an ablation study on depth consistency loss, as shown in Fig.~\ref{fig: ablation-depthloss-2}. We select a room corner, where the input views contain lots of textureless areas. Our depth consistency loss greatly improves the consistency of surface normals and the smoothness of the textureless surfaces.

\section{Conclusion}

We propose NeRFPrior for reconstructing indoor scenes from multi-view images. We introduce to learn a NeRF as a prior which can be trained very fast to sense the geometry and color of a scene. With NeRF prior, we are enabled to use view-dependent color to check visibility, impose multi-view consistency constraints to infer SDF on the surface through volume rendering, and introduce a confidence weighted depth consistency loss to infer planes from textureless areas. Our method provides a novel perspective to learn neural implicit representations from multi-view images through volume rendering, which is much different from the latest methods merely using geometry prior learned in a data-driven or overfitting manner. Our method successfully learns more accurate implicit representations which produces smoother, sharper and more complete surfaces than the state-of-the-art methods. Our experimental results justify the effectiveness and superior of our method.

{
    \small
    \bibliographystyle{ieeenat_fullname}
    \bibliography{main}
}

\end{document}